\documentclass[10pt,a4paper]{article}

\usepackage{ragged2e,float}
\usepackage{graphicx,hyperref}
\usepackage{amsmath,amsthm,amssymb}
\usepackage{bbm}
\usepackage{booktabs,multicol,multirow}
\usepackage[]{todonotes} 
\usepackage{algorithm,algorithmic}
\usepackage[margin=2cm]{geometry}
\usepackage{comment}
\theoremstyle{definition}
\theoremstyle{remark}

\usepackage{soul}

\usepackage{mathrsfs}
\usepackage[numbers,authoryear,sort&compress]{natbib}
\usepackage[T1]{fontenc} 
\usepackage{authblk}
\newcommand{\secref}[1]{\S\ref{#1}}
\newcommand{\etal}{~\textit{et al.}}

\title{An Extensive Evaluation of Factual Consistency in Large Language Models for Data-to-Text Generation}
\author[1]{Joy Mahapatra \thanks{joymahapatra90@gmail.com}}
\author[1]{Utpal Garain \thanks{utpal@isical.ac.in}}
\affil[1]{Indian Statistical Institute Kolkata}

\date{}

\begin{document}
\maketitle

\begin{abstract}
Large Language Models (LLMs) have shown exceptional performance across various Data-to-Text Generation (DTG) tasks.
However, generating factually consistent text in DTG remains challenging for LLMs.
Despite this, in-depth evaluations of LLM factual consistency for DTG remain missing in the current literature.
This paper addresses this gap by providing an extensive evaluation of factual consistency in LLMs for DTG.
Our evaluation covers five widely used DTG datasets (E2E, ViGGo, WikiTableText, DART, and WebNLG) and five prominent LLM families (T5, BART, OPT, BLOOM, and Llama 2). 
To ensure a thorough evaluation of factual consistency, we use four state-of-the-art automatic metrics and include essential human assessments.
Our extensive evaluations reveals three key findings regarding factual consistency in LLMs for DTG.
First, Llama 2 often excels in generating factually consistent text, although smaller models like T5 and BART can achieve strong factual consistency on larger, lexically less-diverse datasets. 
Second, the average rate of change (AROC) indicates that increasing model size (number of model trainable parameters) generally enhances factual consistency of LLMs in DTG.
Third, we observe that source-reference divergence (i.e., when the reference text diverges semantically from the source) typically reduces the factual consistency of LLMs in DTG.
\end{abstract}

\section{Introduction}
\label{sec:introductions}
Data-to-text generation (DTG) aims to generate text, such as reports or dialogues, from structured sources (e.g., tables) or semi-structured sources (e.g., graphs)~\cite{nan2021dart,lin2024survey}.
The DTG task has a wide range of applications across various domains, such as dialogue generation and report creation~\cite{li2024unifying}. 
The rise of large language models (LLMs) has significantly influenced DTG task~\cite{kasner2024reference,lorandi2024high}.  
LLMs have demonstrated considerable improvements over earlier DTG models in several performance areas, including readability (concerns with fluency and coherence) and informativeness (concern with content preservation)~\cite{lin2024survey}.
However, generating factually consistent text for DTG remains a significant challenge for LLMs.
Factual consistency in DTG concerns whether the generated text accurately reflects the factual information from the input/source data~\cite{li2022faithfulness}. 
Lack of factual consistency can severely undermine the trustworthiness of DTG, particularly in safety-critical applications like medical report generation~\cite{yermakov2021biomedical} and financial reporting~\cite{kasner2023tabgenie}.
Hence, evaluation of LLMs' factual consistency for DTG is extremely important.
However, till now there are no extensive evaluation of factual consistency of LLM for DTG.

\begin{figure}[ht]
    \centering
    \includegraphics[width=0.6\linewidth]{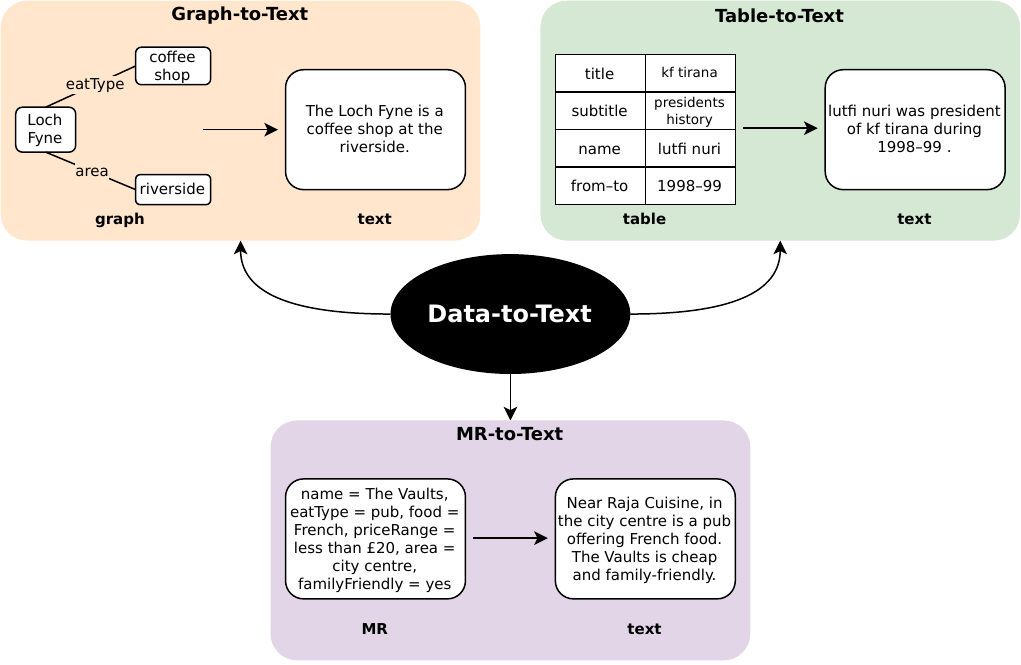}
    \caption{The three primary types of Data-to-Text Generation (DTG): graph-to-text, meaning representation (MR)-to-text, and table-to-text.}
    \label{fig:dtg}
\end{figure}

This paper addresses this research gap by providing an extensive evaluation of the factual consistency of LLMs for DTG through examining multiple datasets and LLM families.
Our study encompasses three primary types of DTG tasks—table-to-text, graph-to-text, and meaning representation (MR)-to-text—as shown in Figure~\ref{fig:dtg}. 
We utilize five state-of-the-art DTG datasets in our experiments: E2E~\cite{dusek2020evaluating} (MR-to-text), ViGGo~\cite{juraska2019viggo} (MR-to-text), WikiTableText~\cite{bao2018table} (table-to-text), DART~\cite{nan2021dart} (graph-to-text), and WebNLG~\cite{gardent2017webnlg} (graph-to-text).
To gain a comprehensive understanding of a wide range of LLMs, we include twelve LLMs of varying sizes from five well-established LLM families: BART~\cite{lewis2020bart}, T5~\cite{raffel2020exploring}, OPT~\cite{zhang2022opt}, BLOOM~\cite{scao2022bloom}, and Llama 2~\cite{touvron2023llama}.
To date, no prior evaluation studies have extensively examined the factual consistency of LLMs for DTG while considering with such large span of LLMs and datasets.
The factual consistency of LLMs is evaluated using both human assessments and automatic metrics, with four distinct automatic metrics: \textsc{SummaC-Conv}, a natural language inference-based metric; \textsc{NEOverlap}, a named entity-based metric; \textsc{AlignScore}, an information alignment-based metric; and \textsc{QAFactEval}, a question generation-answering-based metric.
Moreover, to depict the improvement in factual consistency relative to increasing model size (number of model trainable parameters), we employ the average rate of change (AROC) in our experiments.
Our extensive evaluations and analyses show three key finding on factual consistency of LLMs for DTG.
First, among all LLM families Llama 2 often performs better in producing factually consistent text.
We also observe T5 and BART like small-sized LLM families can produce well factual consistency when dataset size is sufficiency large and contain less lexical diversity (\secref{sec:factual-evalution}).
Secondly, AROC values across various automatic metrics suggests that increasing model size within an LLM family often enhance factual consistency (\secref{sec:factual-evalution}).
The meticulous human evaluation for factual consistency concretized these findings also (\secref{sec:human-evaluation}).
Thirdly, we observe that the presence of source-reference divergence (a common phenomenon in DTG, where the reference text deviates significantly in content from the source data~\cite{dhingra2019handling, li2022faithfulness}) negatively impacts the factual consistency of LLMs in DTG tasks (\secref{sec:results-divergence}).
 
\section{Related Work}
\label{sec:related-work}
Factual consistency is an essential requirement for every DTG model with real-world applications~\cite{gatt2018survey,lin2024survey}. 
Tian\etal~\cite{tian2019sticking} were among the first to identify factual consistency as a critical issue in DTG, referring to it as intrinsic hallucination.
Xiao\etal~\cite{xiao2021hallucination} further explored this issue, linking factual consistency to model predictive uncertainty in DTG tasks.
Additionally, Li\etal~\cite{li2022faithfulness} demonstrated that factual consistency remains a major challenge for DTG models, often surpassing other performance aspects, such as informativeness and readability.
Many other studies~\cite{huang2023survey,li2022faithfulness} across various text generation tasks, including DTG, have attributed factual consistency issues to factors like model architecture, model size, and source-reference divergence.
The emergence of transformer models~\cite{vaswani2017attention}, which rely on full attention mechanisms~\cite{bahdanau2015neural}, has laid the foundation for state-of-the-art LLMs.
Recently, LLMs have been widely adopted as foundational models for DTG tasks due to their vast knowledge capacity enabled by large model sizes, as well as their versatility across various domains, tasks, and languages~\cite{touvron2023llama}.
Although LLMs~\cite{ge2023openagi} often achieve higher rankings than earlier DTG models in readability (addressing concerns with fluency and coherence) and informativeness (maintaining content preservation), they continue to struggle with generating factually consistent content~\cite{ji2023survey,zhang2023sirens}.
Despite several findings on factual consistency challenges in LLMs for DTG~\cite{li2022faithfulness,lin2024survey}, no study has yet conducted a comprehensive evaluation of factual consistency in LLMs. 
While a few studies have explored factual consistency in LLMs for DTG tasks~\cite{lin2024survey}, none have evaluated the factual consistency of state-of-the-art LLMs across diverse DTG tasks.
Our paper aims to address this critical research gap by providing an extensive evaluation of factual consistency, incorporating state-of-the-art LLMs and a range of DTG datasets.

\section{Preliminaries}
In this preliminaries section, we provide a brief overview of data-to-text generation (DTG), large language models, factual consistency, and source-reference divergence.

\subsection{Data-to-Text Generation (DTG)}
\label{subsec:dtg}
Data-to-Text generation (DTG)~\cite{lin2024survey} model is usually learned from a dataset consisting of a set of source-reference pairs, $\mathcal{D} = \left.{\{s, r\}}\right \vert_{\in \mathcal{D}}$, where $s$ is source data (e.g., table, graph, etc.) with its corresponding reference $r = r_{1}r_{2},\dots,r_{|r|}$, and each word belong to vocabulary set $\mathcal{V}$, i.e., $r_{i} \in \mathcal{V}$~\cite{puduppully2021data}.
A predictive DTG model $\mathcal{M}_{\theta}$ ($\theta$ is the parameter set) is trained/fine-tuned on $\mathcal{D}$, typically using maximum likelihood estimation~\cite{puduppully2021data} strategies (\autoref{eq:training}).

\begin{align}
    \theta^{*} \leftarrow &\underset{\theta}{\text{argmax}}\sum\limits_{(s,r) \in \mathcal{D}}\mathcal{L}(\mathcal{M}_\theta(s), r) \label{eq:training}
\end{align}

Here, $\mathcal{L}$ represents empirical loss function (e.g., cross-entropy loss).
Nowadays, LLMs are often used as predictive DTG models.
During inference, trained predictive DTG model, $\mathcal{M}_{\theta^*}$, undergoes through applying various decoding strategies~\cite{li2022faithfulness} to produce output $g$ (an approximation of $r$) from source data ($s$).

\begin{align}
    g \leftarrow \underset{\mathcal{V}^{*}}{\text{decoding}}\ \mathcal{M}_{\theta^*}(s) \label{eq:decoding}
\end{align}

\subsection{(Large) Language Model}
\label{subsec:llm}
The core objective of a language model is to predict text continuation ($y = w_1w_2 \dots w_{m}$) from a given prior contexts ($x=w_1w_2 \dots w_{n}$), as shown below.
\begin{align*}
    y \leftarrow & \underset{z}{\text{argmax}}\ p(z \vert x)\\
    \underbrace{w_1w_2 \dots w_{m}}_{\text{text continuation}} \leftarrow & \underset{z}{\text{argmax}}\ p(z \vert \underbrace{w_1w_2 \dots w_{n}}_{\text{context}})
\end{align*}

\begin{figure}[ht]
    \centering
    \resizebox{0.6\linewidth}{!}{
    \includegraphics{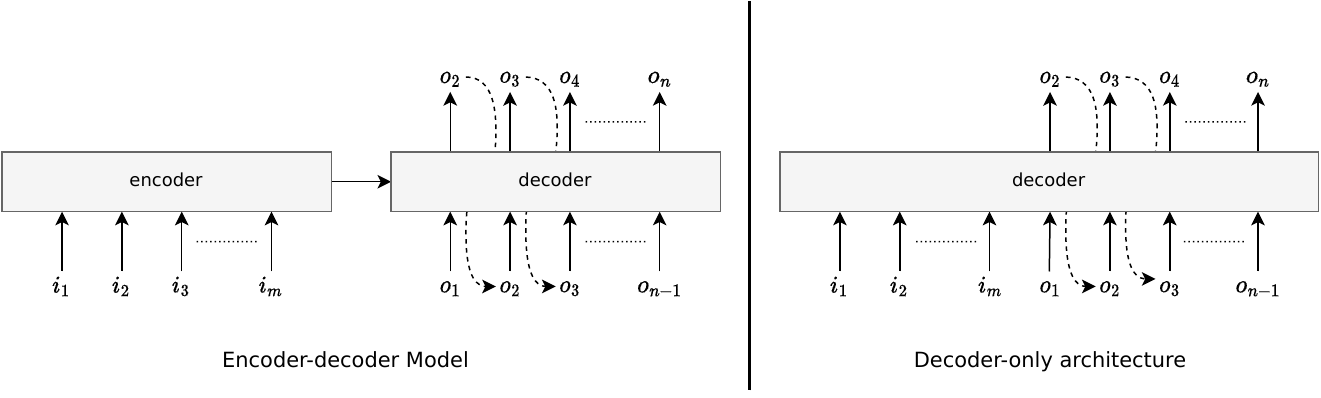}}
    \caption{Two popular language model architectures: encoder-decoder (left) and decoder-only (right). The encoder processes input ($i$); the decoder generates output ($o$).}
    \label{fig:seq2seq}
\end{figure}

Two popular architectures are commonly used for implementing language models: encoder-decoder architecture~\cite{lewis2020bart} and decoder-only architecture~\cite{brown2020language} (Figure~\ref{fig:seq2seq}).
``Large" language models (LLMs) stand out from previous language models mainly due to two aspects: their enormous model size~\cite{scao2022bloom}, characterized by a vast number of underlying parameters, and their extensive pre-training data~\cite{raffel2020exploring,penedo2023refinedweb}.
Compared to earlier language models with a few million parameters~\cite{graves2013speech,mikolov2010recurrent,peters2018deep}, contemporary LLMs have billions, and even trillions, of parameters, enabling them to achieve exceptional performance~\cite{heinzerling2021language}.

\begin{figure}[h]
    \centering
    \includegraphics[width=0.6\linewidth]{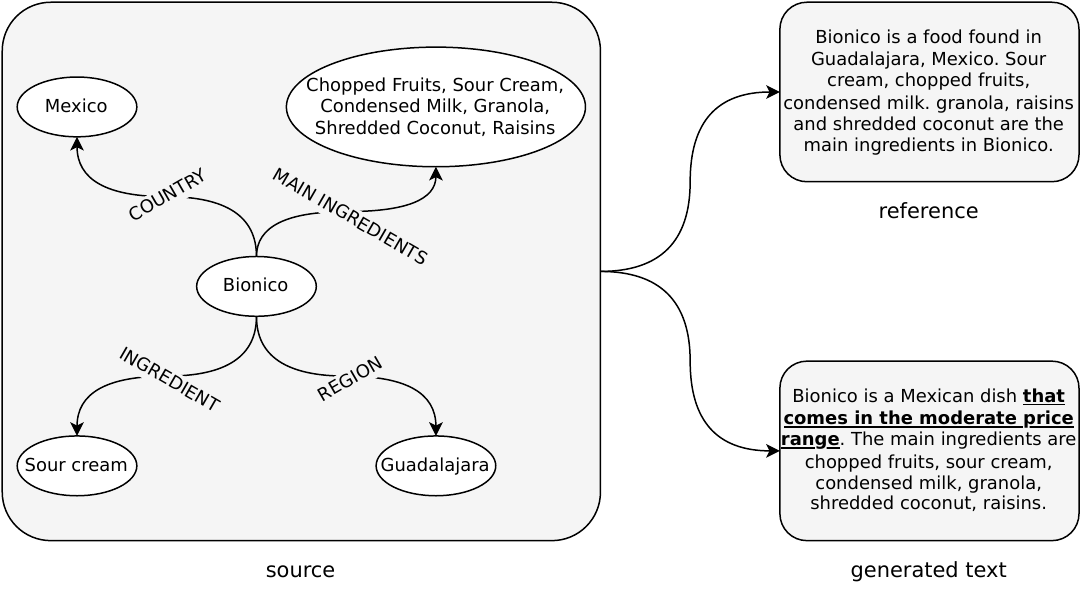}
    \caption{An example of lack of factual consistency from the DART (graph-to-text) dataset, where the claim (generated text) introduces incorrect information—\textbf{\ul{`that comes in the moderate price range'}}—not found in the source (reference) data.} 
    \label{fig:example-factual-consistency}
\end{figure}

\subsection{Factual Consistency in DTG}
\label{subsec:factual-consistency-dtg}
Factual consistency is typically assessed by comparing the `claim' against the `context' to ensure factual alignment. 
In DTG, the reference text (or sometimes source data) serves as the context, while the generated text acts as the claim. 
As illustrated in~\autoref{fig:example-factual-consistency}, a lack of factual consistency can lead to the introduction of irrelevant or incorrect information, undermining the model's reliability.
Evaluating a DTG model's factual consistency is therefore crucial for determining its reliability.
The most common methods for assessing factual consistency are human evaluation and automatic metric-based evaluation.
Recently, some trained automatic metrics have shown strong correlations with human assessments, suggesting that they can be reliable indicators of factual consistency.

\subsection{Source-Reference Divergence}
\label{subsec:divergence}
\begin{figure}[ht]
    \centering
    \resizebox{0.6\linewidth}{!}{
    \includegraphics{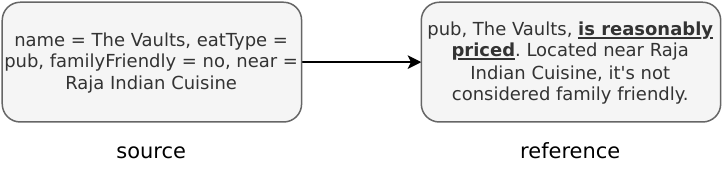}}
    \caption{An illustration of source-reference divergence from E2E~\cite{dusek2020evaluating} dataset. 
    The reference text includes an additional facts (bolded and underlined)---\textbf{\ul{`is reasonably priced'}}---which in absent in the source data.}
    \label{fig:divergence}
\end{figure}

Source-reference divergence is an extremely common phenomenon across various DTG tasks~\cite{dhingra2019handling,tian2019sticking,islam2023tackling,li2022faithfulness}.
Source-reference divergence refers to a discrepancy or divergence between the information present in the source data and the corresponding reference data. 
An illustrative example of such divergence is depicted in Figure~\ref{fig:divergence}. 
The main causes of source-reference divergence stem from the inherent characteristics of DTG tasks and heuristic data collection procedures~\cite{li2022faithfulness}.
As a result, source-reference divergence is common and difficult to eliminate in DTG contexts~\cite{dhingra2019handling}. 
Therefore, evaluating model factual consistency in the presence of such divergence is essential.

\section{Experimental Setup}
\label{sec:experimental-setup}
This section describes the datasets, LLMs, and experimental settings used in our experiments, along with statistical significance tests to ensure reproducibility.

\subsection{Datasets}
We use five well-known DTG datasets for three primary task types: DART and WebNLG for graph-to-text, WikiTableText for table-to-text, and E2E and ViGGO for MR-to-text.
The key statistics for these datasets are in Table~\ref{tab:dataset_statistics}, all obtained from \cite{kasner2023tabgenie} and \cite{wolf2020transformers}. 
E2E~\cite{dusek2020evaluating} is a restaurant-domain MR-to-text dataset, and ViGGO~\cite{juraska2018deep} focuses on dialogue MR-to-text in the video game domain. 
WikiTableText~\cite{bao2018table} provides table-to-text pairs from Wikipedia with high lexical diversity (or lexical richness) (as TTR is higher for both source and reference).
DART~\cite{nan2021dart} provides text from knowledge graph triplets, while WebNLG~\cite{gardent2017webnlg} focuses on RDF-to-text from DBPedia triplets.

\begin{table*}[ht]
    \centering
    \resizebox{\textwidth}{!}{
    \begin{tabular}{@{}cccccccccc@{}}
    \toprule
    \multirow{2}{*}{dataset} & \multirow{2}{*}{DTG types} & \multirow{2}{*}{domain} & size & \multicolumn{3}{c}{source (linearized)} & \multicolumn{3}{c}{reference} \\ \cmidrule(lr){5-7} \cmidrule(lr){8-10} 
     &  &  & (\# instances) & type-token ratio (TTR) & unique tokens & total tokens & type-token ratio (TTR) & unique tokens & total tokens \\ \midrule
    E2E & mr-to-text & closed & 36,856 & 0.0001 & 125 & 1M & 0.0057 & 4.5K & 885K\\
    ViGGo & mr-to-text & closed & 6,900 & 0.0029 & 618 & 206K & 0.0294 & 4.4K & 148K \\ \midrule
    WikiTableText & table-to-text & open & 13,318 & 0.0621 & 29K & 469K & 0.1302 & 24K & 185K \\ \midrule
    DART & graph-to-text & open & 70,524 & 0.0178 & 44K & 2.5M & 0.0328 & 45K & 1.5M \\
    WebNLG & graph-to-text & open & 38,872 & 0.0057 & 7K & 1.2M & 0.0228 & 19K & 905K \\ \bottomrule
    \end{tabular}}
    \caption{Summary of key statistics, DTG types, and domains for the five incorporated datasets. 
    The table shows the average length (in tokens), unique tokens, and total tokens for both sources (linearized to text) and references.}
    \label{tab:dataset_statistics}
\end{table*}

\subsection{Models}
\begin{figure}[h]
    \centering
    \resizebox{0.6\linewidth}{!}{
    \includegraphics[]{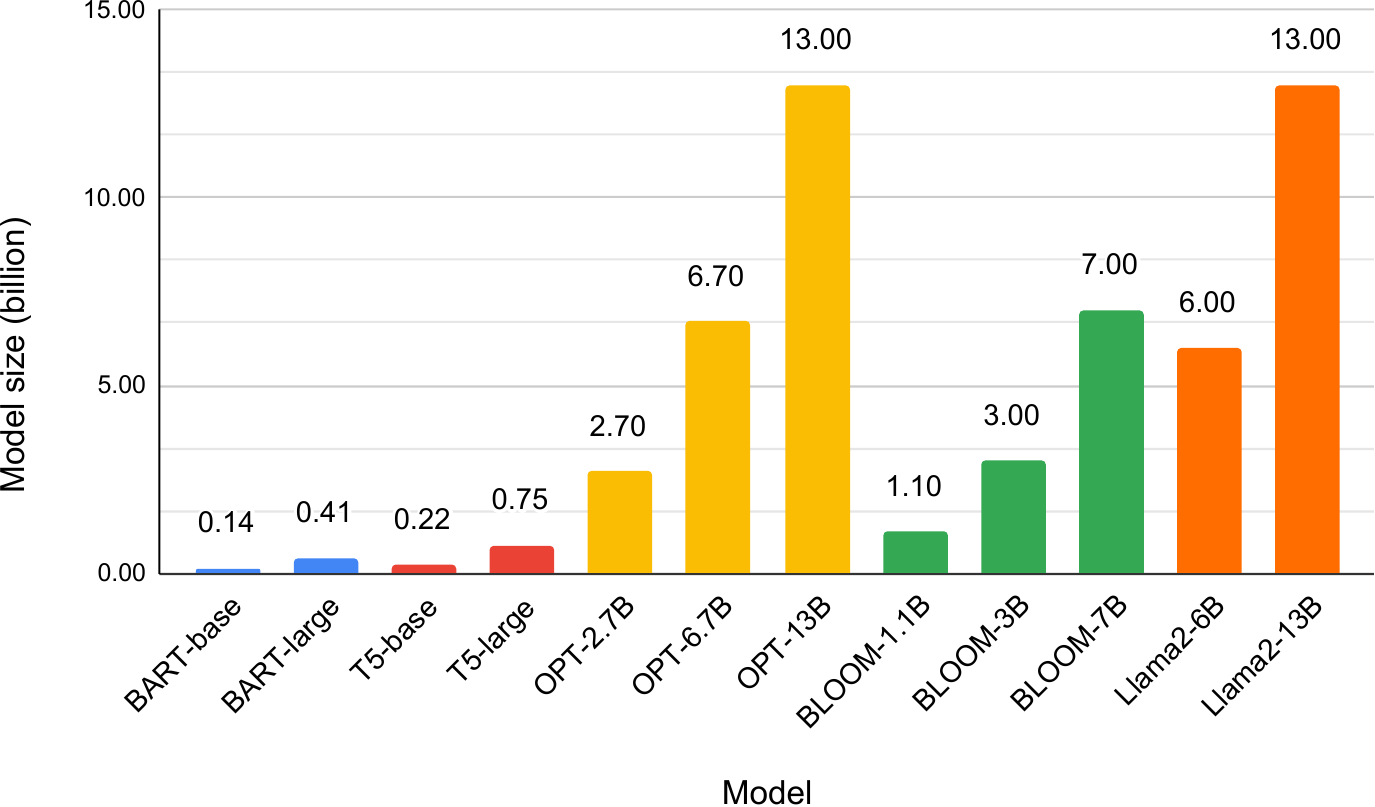}}
    \caption{All twelve LLMs across five widely used families (BART, T5, OPT, BLOOM, and Llama2) arranged by their model sizes.}
    \label{fig:12-llms}
\end{figure}

We include twelve LLMs from five families—BART, T5, BLOOM, OPT, and Llama 2—downloaded from Hugging Face~\cite{wolf2020transformers}. 
All these twelve LLMs are shown in Figure~\ref{fig:12-llms}, along with their sizes.
For the BART, T5, OPT, BLOOM, and Llama 2 families, we select two models from BART and T5, three models from OPT and BLOOM, and two models from Llama 2.
BART~\cite{lewis2020bart} and T5~\cite{raffel2020exploring} are encoder-decoder models, with BART pre-trained using text denoising and T5 with unified NLP tasks.
BLOOM~\cite{scao2022bloom} and OPT~\cite{zhang2022opt} are decoder-only models, trained on ROOT and Reddit/Pile/RoBERTa, respectively. 
Llama 2~\cite{touvron2023llama} is trained with pre-training, fine-tuning, and reinforcement learning. 
These LLM families were selected for their varied pretraining dataset and sizes, enabling a comprehensive evaluation of factual consistency across different DTG tasks.

\subsection{Experimental Setting and Statistical Significant Testing Details}
\label{subsec:setting-significance}
In our experiments, we fine-tuned all LLMs for DTG tasks using QLoRA (Quantized Low-Rank Adapter)~\cite{dettmers2023qlora}, a parameter-efficient fine-tuning method with a learning rate of $1e^{-04}$ and 4-bit quantization.
Each model was fine-tuned for three epochs on all five datasets, with source and generated text lengths fixed at 256 tokens.
We used the AdamW optimizer~\cite{loshchilov2019decoupled} at \texttt{torch.bfloat16} precision for QLoRA.
Text generation was performed through popular nucleus sampling decoding approach with nucleus size ($p=0.95$).
All experiments were run on a local system with an NVIDIA RTX A6000 (48 GB) GPU.
We assessed statistical significance using Welch's t-test~\cite{dror2018hitchhikers}, comparing each LLM to the best performer in its family, with a significance level of $p < 0.05$ and a sample size of 6.

\subsection{Average Rate of Change (AROC)}
Increasing model size within an LLM family often affects factual consistency. This effect can be quantified using the average rate of change (AROC), which estimates the change in factual consistency between the smallest model $\mathcal{M}_1$ and the largest model $\mathcal{M}_2$ of the family, based on their sizes $m_1$ and $m_2$. AROC is calculated within the range $[m_1, m_2]$ using~\autoref{eq:aroc}.

\begin{align}
    \text{AROC} &= \frac{\text{absolute change in factual consistency}}{\text{absolute change in model size}} \nonumber\\
    &= \frac{f(\mathcal{M}_2)-f(\mathcal{M}_1)}{m_2 - m_1} \label{eq:aroc}
\end{align}

\begin{figure}[h]
    \centering
    \resizebox{0.4\linewidth}{!}{
    \includegraphics[]{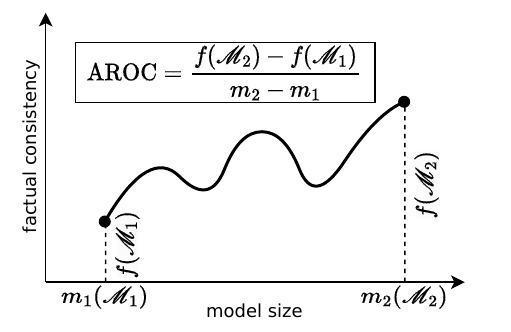}}
    \caption{AROC of the function $f(\cdot)$ (representing factual consistency) for the model size interval $[m_1, m_2]$ between two LLMs, $\mathcal{M}_1$ and $\mathcal{M}_2$.}
    \label{fig:aroc}
\end{figure}

LLM families differ significantly in model size scales; for instance, T5 sizes are in millions, while Llama2 sizes are in billions, impacting factual consistency. Comparing LLM families using AROC (\autoref{eq:aroc}) alone is therefore insufficient. 
Instead, we apply AROC on a logarithmic scale (\autoref{eq:aroc-log}), capturing relative changes by converting absolute differences into ratios (\autoref{eq:aroc-log-ratio}).

\begin{align}
    \text{AROC} &= \frac{\operatorname{log}(f(\mathcal{M}_2))-\operatorname{log}(f(\mathcal{M}_1))}{\operatorname{log}(m_2) - \operatorname{log}(m_1)} \label{eq:aroc-log}\\
    &= \frac{\operatorname{log}(f(\mathcal{M}_2)/f(\mathcal{M}_1))}{\operatorname{log}(m_2/m_1)} \label{eq:aroc-log-ratio}
\end{align}

A positive AROC value (\autoref{eq:aroc-log} or~\ref{eq:aroc-log-ratio}) indicates a positive correlation between model size increment and factual consistency, while the magnitude of the AROC value reflects the strength of this correlation.

\section{Evaluation of Factual Consistency: Results and Discussion}
\label{sec:factual-evalution}
This section presents our extensive evaluation of factual consistency across five widely used LLM families and five prominent DTG datasets, employing four state-of-the-art metrics: \textsc{SummaC-Conv}, \textsc{NEOverlap}, \textsc{AlignScore}, and \textsc{QAFactEval}.

\begin{table}[h]
    \centering
    \caption{Factual consistency evaluated through \textsc{SummaC-Conv}. 
    \textbf{Boldface} shows best result per LLM family and dataset.
    All results statistically significant (see~\ref{subsec:setting-significance}).}
    \label{tab:summac-conv-numerical}
    \resizebox{0.6\linewidth}{!}{
    \begin{tabular}{@{}cccccccc@{}}
        \toprule
        family & model & model size (billion) & E2E & WikiTableText & ViGGo & DART & WebNLG \\ \midrule
        \multirow{2}{*}{BART} & BART-base & 0.14 & 0.3805 & 0.3111 & 0.2330 & 0.3004 & 0.2894 \\
         & BART-large & 0.41 & \textbf{0.4110} & \textbf{0.3609} & \textbf{0.2498} & \textbf{0.3786} & \textbf{0.4173} \\ \midrule
        \multirow{2}{*}{T5} & T5-base & 0.22 & 0.3899 & 0.3625 & 0.2435 & 0.3582 & 0.3687 \\
         & T5-large & 0.75 & \textbf{0.4135} & \textbf{0.4082} & \textbf{0.2684} & \textbf{0.4264} & \textbf{0.4593} \\ \midrule
        \multirow{3}{*}{OPT} & OPT-2.7B & 2.70 & \textbf{0.3637} & \textbf{0.4120} & 0.2971 & 0.3926 & 0.4781 \\
         & OPT-6.7B & 6.70 & 0.3597 & 0.4060 & 0.3139 & 0.3937 & 0.4799 \\
         & OPT-13B & 13.00 & 0.3621 & 0.4092 & \textbf{0.3291} & \textbf{0.4137} & \textbf{0.4970} \\ \midrule
        \multirow{3}{*}{BLOOM} & BLOOM-1.1B & 1.10 & 0.3777 & 0.3871 & 0.2874 & 0.3850 & 0.3853 \\
         & BLOOM-3B & 3.00 & 0.3842 & 0.4136 & \textbf{0.3177} & 0.4119 & 0.4610 \\
         & BLOOM-7B & 7.00 & \textbf{0.3872} & \textbf{0.4184} & 0.3170 & \textbf{0.4144} & \textbf{0.4780} \\ \midrule
        \multirow{2}{*}{Llama 2} & Llama2-7B & 6.00 & \textbf{0.4569} & 0.4898 & 0.3622 & 0.5138 & \textbf{0.5739} \\
         & Llama2-13B & 13.00 & 0.4540 & \textbf{0.5009} & \textbf{0.3776} & \textbf{0.5201} & 0.5693 \\ \bottomrule
    \end{tabular}}
\end{table}

\begin{figure}[h]
    \centering
    \resizebox{0.6\linewidth}{!}{
    \includegraphics[]{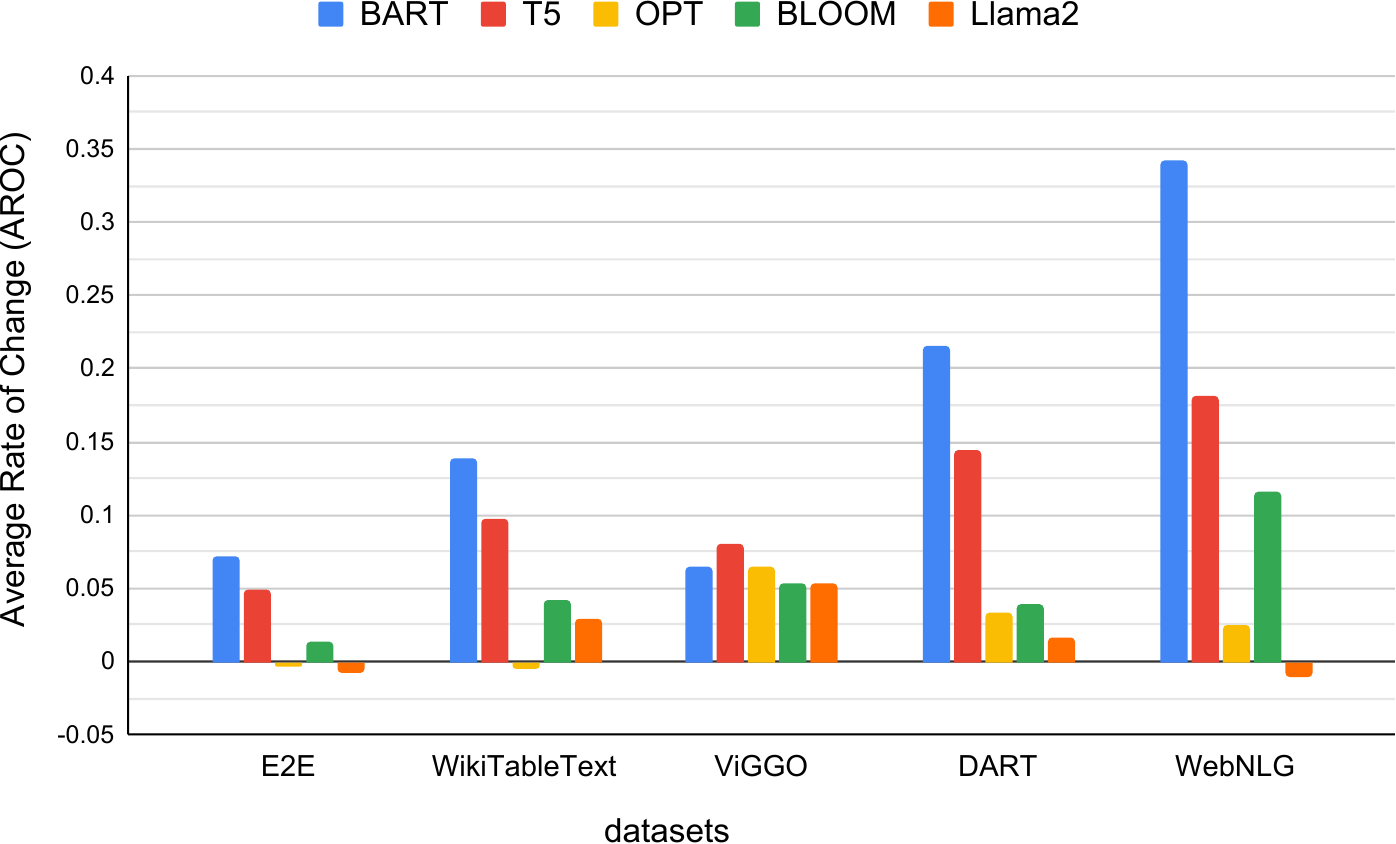}}
    \caption{AROC values for all LLM families across the five DTG datasets based on \textsc{SummaC-Conv} scores.
    Higher positive AROC values reflect a stronger positive correlation between increases in model size and factual consistency.}
    \label{fig:summac-conv-aroc}
\end{figure}

\subsection{Results of Factual Consistency Evaluation Using \textsc{SummaC-Conv}}
\label{subsec:summac-conv}
NLI (natural language inference) models are commonly used to assess factual consistency due to their similarity to the task.
However, a key limitation of NLI models is their focus on sentence-level granularity, which is insufficient for evaluating factual consistency that requires document-level analysis.
\textsc{SummaC-Conv}~\cite{laban2022summac} addresses this limitation by providing document-level granularity.
It segments both the context and claim into sentences, computing NLI scores for each sentence pair.
Additionally, \textsc{SummaC-Conv} employs binning techniques and one-dimensional convolution to mitigate the impact of outliers, effectively capturing the factual consistency between the claim and context.
\textsc{SummaC-Conv} is widely regarded as a highly popular metric for evaluating factual consistency.

\autoref{tab:summac-conv-numerical} presents the factual consistency scores measured by \textsc{SummaC-Conv} across the five LLM families and five DTG datasets.
A higher \textsc{SummaC-Conv} value indicates better factual consistency.
Meanwhile, \autoref{fig:summac-conv-aroc} displays the AROC values of \textsc{SummaC-Conv}, derived from \autoref{tab:summac-conv-numerical}, for all LLM families and datasets.

\begin{table}[h]
    \centering
    \caption{Factual consistency evaluated through \textsc{NEOverlap}. 
    \textbf{Boldface} shows best result per LLM family and dataset.
    All results statistically significant (see~\ref{subsec:setting-significance}).}
    \label{tab:neoverlap-numerical}
    \resizebox{0.6\linewidth}{!}{
    \begin{tabular}{@{}cccccccc@{}}
        \toprule
        family & model & model size (billion) & E2E & WikiTableText & ViGGo & DART & WebNLG \\ \midrule
        \multirow{2}{*}{BART} & BART-base & 0.14 & 0.8305 & 0.5183 & 0.7077 & 0.5643 & 0.3725 \\
         & BART-large & 0.41 & \textbf{0.8368} & \textbf{0.5700} & \textbf{0.7892} & \textbf{0.6204} & \textbf{0.4761} \\ \midrule
        \multirow{2}{*}{T5} & T5-base & 0.22 & 0.8305 & 0.5556 & 0.7815 & 0.6164 & 0.4195 \\
         & T5-large & 0.75 & \textbf{0.8593} & \textbf{0.5982} & \textbf{0.8985} & \textbf{0.6813} & \textbf{0.5021} \\ \midrule
        \multirow{3}{*}{OPT} & OPT-2.7B & 2.70 & 0.8323 & \textbf{0.6267} & 0.8908 & 0.6181 & 0.5385 \\
         & OPT-6.7B & 6.70 & 0.8142 & 0.6058 & 0.9215 & 0.6358 & 0.5546 \\
         & OPT-13B & 13.00 & \textbf{0.8431} & 0.6075 & \textbf{0.9369} & \textbf{0.6499} & \textbf{0.5546} \\ \midrule
        \multirow{3}{*}{BLOOM} & BLOOM-1.1B & 1.10 & 0.8133 & 0.5858 & 0.8938 & 0.5973 & 0.4373 \\
         & BLOOM-3B & 3.00 & 0.8070 & 0.6050 & 0.9200 & \textbf{0.6326} & 0.4844 \\
         & BLOOM-7B & 7.00 & \textbf{0.8206} & \textbf{0.6208} & \textbf{0.9200} & 0.6310 & \textbf{0.5203} \\ \midrule
        \multirow{2}{*}{Llama 2} & Llama2-7B & 6.00 & 0.8674 & 0.6350 & 0.9462 & \textbf{0.7269} & 0.6368 \\
         & Llama2-13B & 13.00 & \textbf{0.8683} & \textbf{0.6675} & \textbf{0.9523} & 0.7158 & \textbf{0.6532} \\ \bottomrule
    \end{tabular}}
\end{table}

\begin{figure}[h!]
    \centering
    \resizebox{0.6\linewidth}{!}{
    \includegraphics[]{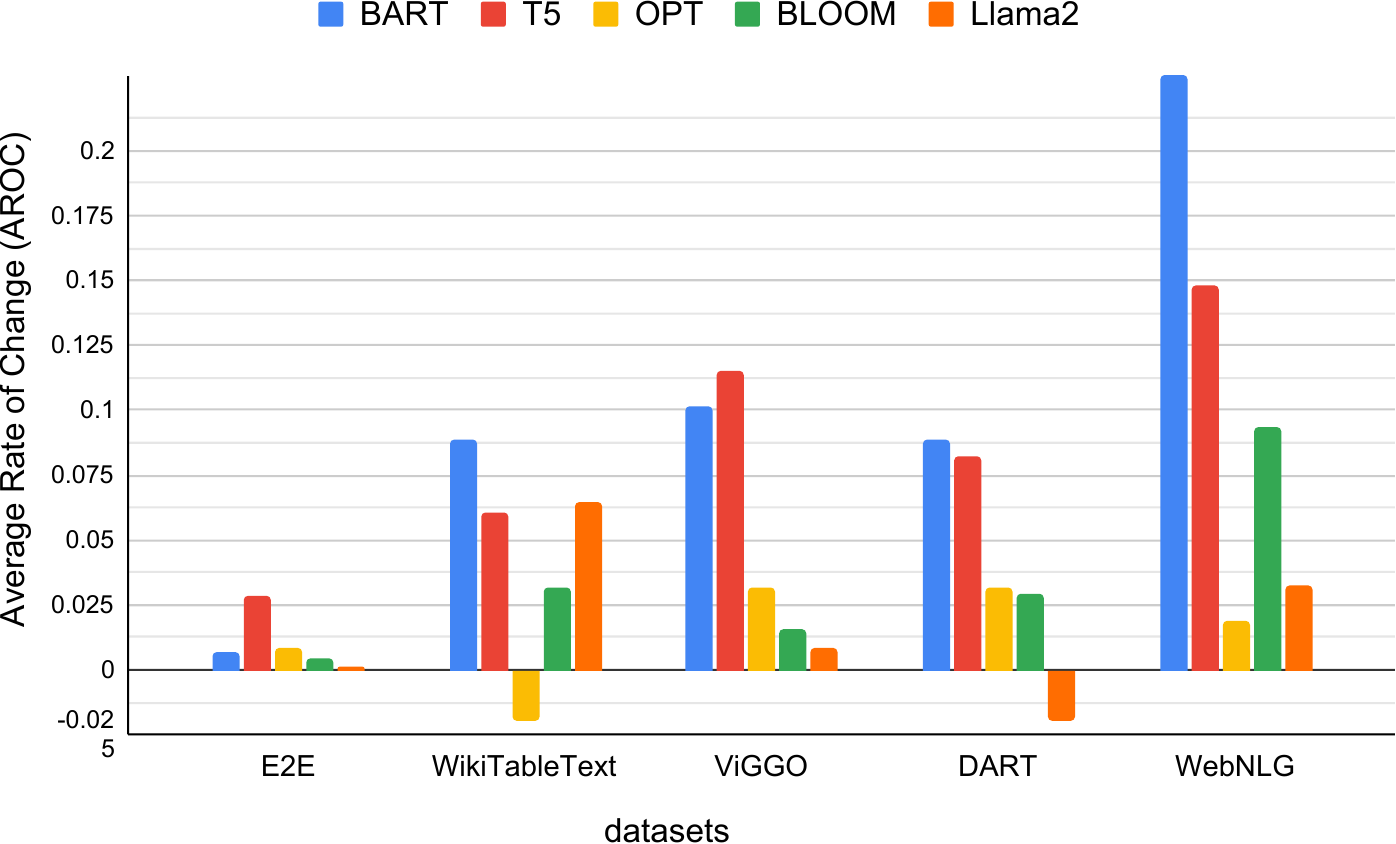}}
    \caption{AROC values for all LLM families across the five DTG datasets based on \textsc{NEOverlap} scores.
    Higher positive AROC values reflect a stronger positive correlation between increases in model size and factual consistency.}
    \label{fig:neoverlap-aroc}
\end{figure}

\subsection{Results of Factual Consistency Evaluation Using \textsc{NEOverlap}}
\label{subsec:neoverlap}
Named entities (NEs) are key indicators of factual knowledge in texts and are often used to assess factual consistencies.
\textsc{NEOverlap} measures factual consistency by comparing the overlap of NEs between the claim and the context.
We utilized the metric implementation from~\cite{laban2022summac}, focusing on specific NE types such as \texttt{PERSON}, \texttt{LOCATION}, and \texttt{ORGANIZATION}.
This metric returns a value of 1 if the generated text is factually consistent with the reference in terms of NE overlap, and 0 if there is little or no factual consistency.
In the DTG context, the reference text is treated as the context, and the generated text as the claim.

\autoref{tab:neoverlap-numerical} displays the factual consistency scores evaluated using \textsc{NEOverlap} across the five LLM families and DTG datasets. 
~\autoref{fig:neoverlap-aroc} presents the AROC values of \textsc{NEOverlap}, derived from \autoref{tab:neoverlap-numerical}, for all LLM families and datasets.

\begin{table}[h]
    \centering
    \caption{Factual consistency evaluated through \textsc{AlignScore}. 
    \textbf{Boldface} shows best result per LLM family and dataset.
    All results statistically significant (see~\ref{subsec:setting-significance}).}
    \label{tab:alignscore-numerical}
    \resizebox{0.6\linewidth}{!}{
    \begin{tabular}{@{}cccccccc@{}}
        \toprule
        family & model & model size (billion) & E2E & WikiTableText & ViGGo & DART & WebNLG \\ \midrule
        \multirow{2}{*}{BART} & BART-base & 0.14 & 0.6825 & 0.3189 & 0.2579 & 0.4064 & 0.3869 \\
        & BART-large & 0.41 & \textbf{0.7663} & \textbf{0.4158} & \textbf{0.3538} & \textbf{0.6013} & \textbf{0.6551} \\ \midrule
        \multirow{2}{*}{T5} & T5-base & 0.22 & 0.7049 & 0.4045 & 0.3238 & 0.5755 & 0.5457 \\
        & T5-large & 0.75 & \textbf{0.7646} & \textbf{0.4513} & \textbf{0.4459} & \textbf{0.7050} & \textbf{0.7207} \\ \midrule
        \multirow{3}{*}{OPT} & OPT-2.7B & 2.70 & 0.6931 & \textbf{0.4897} & 0.5439 & 0.6326 & 0.7764 \\
        & OPT-6.7B & 6.70 & 0.6792 & 0.4835 & 0.6036 & 0.6374 & 0.7876 \\
        & OPT-13B & 13.00 & \textbf{0.6999} & 0.4810 & \textbf{0.6356} & \textbf{0.6714} & \textbf{0.8145} \\ \midrule
        \multirow{3}{*}{BLOOM} & BLOOM-1.1B & 1.10 & 0.7147 & 0.4887 & 0.4548 & 0.6175 & 0.6099 \\
        & BLOOM-3B & 3.00 & \textbf{0.7468} & \textbf{0.5326} & 0.5653 & 0.6723 & 0.7507 \\
        & BLOOM-7B & 7.00 & 0.7280 & 0.5296 & \textbf{0.5884} & \textbf{0.6848} & \textbf{0.7797} \\ \midrule
        \multirow{2}{*}{Llama 2} & Llama2-7B & 6.00 & 0.8106 & 0.5905 & 0.6798 & \textbf{0.8192} & \textbf{0.8768} \\
        & Llama2-13B & 13.00 & \textbf{0.8186} & \textbf{0.6144} & \textbf{0.7163} & 0.8191 & 0.8749 \\ \bottomrule
    \end{tabular}}
\end{table}

\begin{figure}[h!]
    \centering
    \resizebox{0.6\linewidth}{!}{
    \includegraphics[]{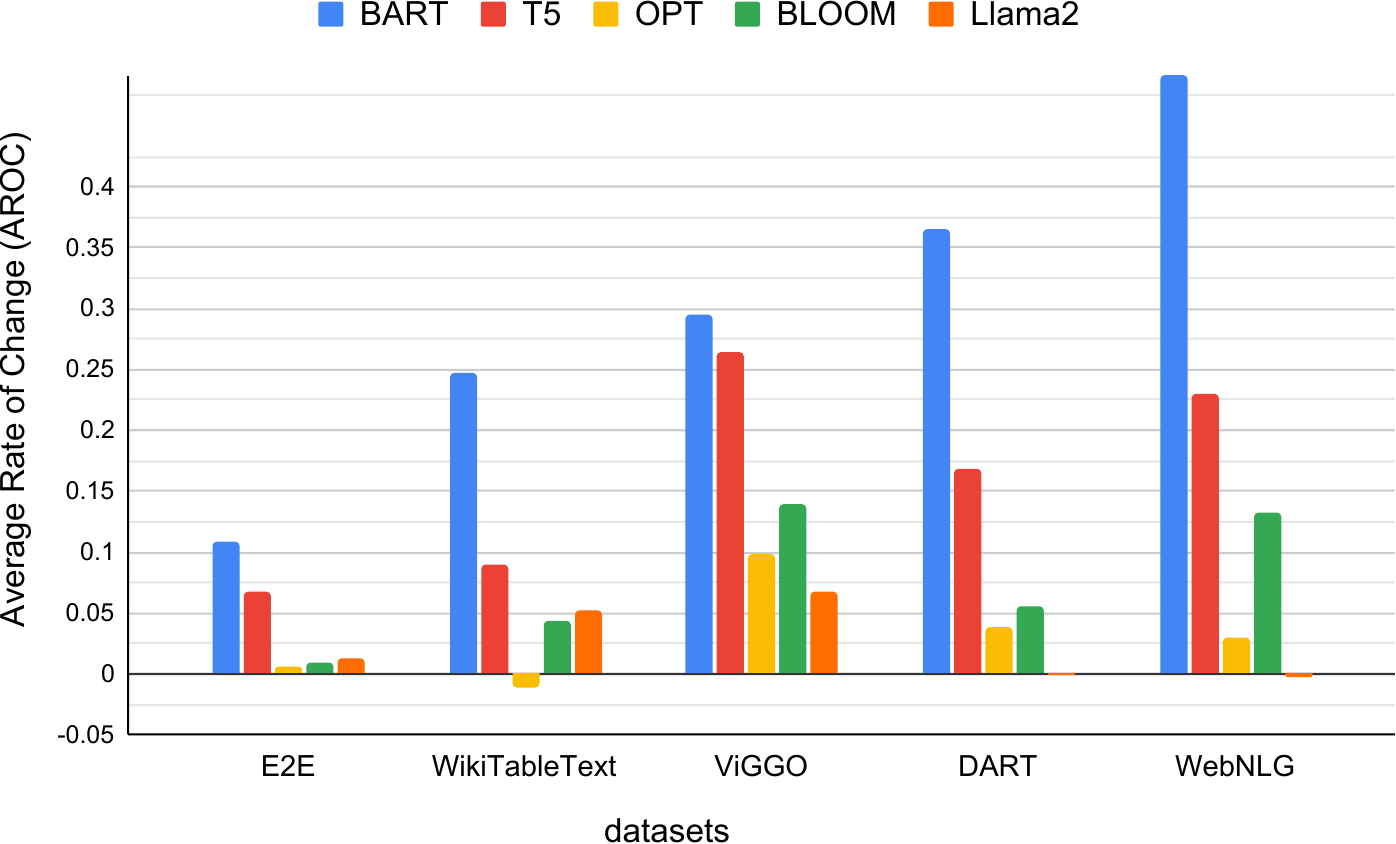}}
    \caption{AROC values for all LLM families across the five DTG datasets based on \textsc{AlignScore} scores.
    Higher positive AROC values reflect a stronger positive correlation between increases in model size and factual consistency.}
    \label{fig:alignscore-aroc}
\end{figure}

\subsection{Results of Factual Consistency Evaluation Using \textsc{AlignScore}}
\label{subsec:alignscore-result}
Unlike most factual consistency metrics, which typically focus on single tasks like semantic similarity or NLI (natural language inference), \textsc{AlignScore}~\cite{zha2023alignscore} is trained on multiple tasks (fact verification, information retrieval, NLI, QA, paraphrasing, semantic similarity, and summarization) in a unified manner. 
\textsc{AlignScore} assesses factual consistency by analyzing the alignment of information between the claim and context, capturing various qualities, including factual consistency.
A key feature of \textsc{AlignScore} is its ability to handle longer contexts by splitting the context into chunks and aligning each chunk with the claim's sentences.
A higher \textsc{AlignScore} reflects greater factual consistency.
In our experiment, \textsc{AlignScore} refers specifically to the \textsc{AlignScore-large} variant, which utilizes the \textsf{RoBERTa-large} model for information alignment.
Zha\etal~\cite{zha2023alignscore} shows that \textsc{AlignScore} outperforms various state-of-the-art measures, including LLM-based methods, in evaluating factual accuracy.

\autoref{tab:alignscore-numerical} presents the factual consistency scores evaluated using \textsc{AlignScore} across all LLM families and DTG datasets. 
~\autoref{fig:alignscore-aroc} shows the AROC values of \textsc{AlignScore}, obtained from \autoref{tab:alignscore-numerical}, for all LLM families and datasets.

\begin{table}[h]
    \centering
    \caption{Factual consistency evaluated through \textsc{QAFactEval}. 
    \textbf{Boldface} shows best result per LLM family and dataset.
    All results statistically significant (see~\ref{subsec:setting-significance}).}
    \label{tab:qafacteval-numerical}
    \resizebox{0.6\linewidth}{!}{
    \begin{tabular}{@{}cccccccc@{}}
        \toprule
        family & model & model size (billion) & E2E & WikiTableText & ViGGo & DART & WebNLG \\ \midrule
        \multirow{2}{*}{BART} & BART-base & 0.14 & 3.4109 & 1.8008 & 1.3894 & 2.2541 & 2.1009 \\
         & BART-large & 0.41 & \textbf{3.7393} & \textbf{2.3559} & \textbf{1.9500} & \textbf{3.1206} & \textbf{3.3009} \\ \midrule
        \multirow{2}{*}{T5} & T5-base & 0.22 & 3.6086 & 2.1887 & 1.6307 & 2.7631 & 2.5079 \\
         & T5-large & 0.75 & \textbf{3.7393} & \textbf{2.6267} & \textbf{2.0971} & \textbf{3.4325} & \textbf{3.3865} \\ \midrule
        \multirow{3}{*}{OPT} & OPT-2.7B & 2.70 & 3.2732 & 2.5878 & 2.4776 & 3.1245 & 3.5771 \\
         & OPT-6.7B & 6.70 & \textbf{3.2930} & \textbf{2.6805} & 2.7332 & 3.1335 & 3.6794 \\
         & OPT-13B & 13.00 & 3.2768 & 2.6085 & \textbf{2.8302} & \textbf{3.2894} & \textbf{3.7640} \\ \midrule
        \multirow{3}{*}{BLOOM} & BLOOM-1.1B & 1.10 & 3.5579 & 2.2128 & 2.0289 & 3.0540 & 2.9072 \\
         & BLOOM-3B & 3.00 & 3.6179 & 2.4706 & 2.5181 & 3.2751 & 3.4617 \\
         & BLOOM-7B & 7.00 & \textbf{3.6245} & \textbf{2.5299} & \textbf{2.5265} & \textbf{3.3069} & \textbf{3.5181} \\ \midrule
        \multirow{2}{*}{Llama 2} & Llama2-7B & 6.00 & 4.1619 & 3.0296 & 2.9628 & 4.0208 & \textbf{4.1166} \\
         & Llama2-13B & 13.00 & \textbf{4.1874} & \textbf{3.0933} & \textbf{3.0950} & \textbf{4.0411} & 4.0892 \\ \bottomrule
    \end{tabular}}
\end{table}

\begin{figure}[h]
    \centering
    \resizebox{0.6\linewidth}{!}{
    \includegraphics[]{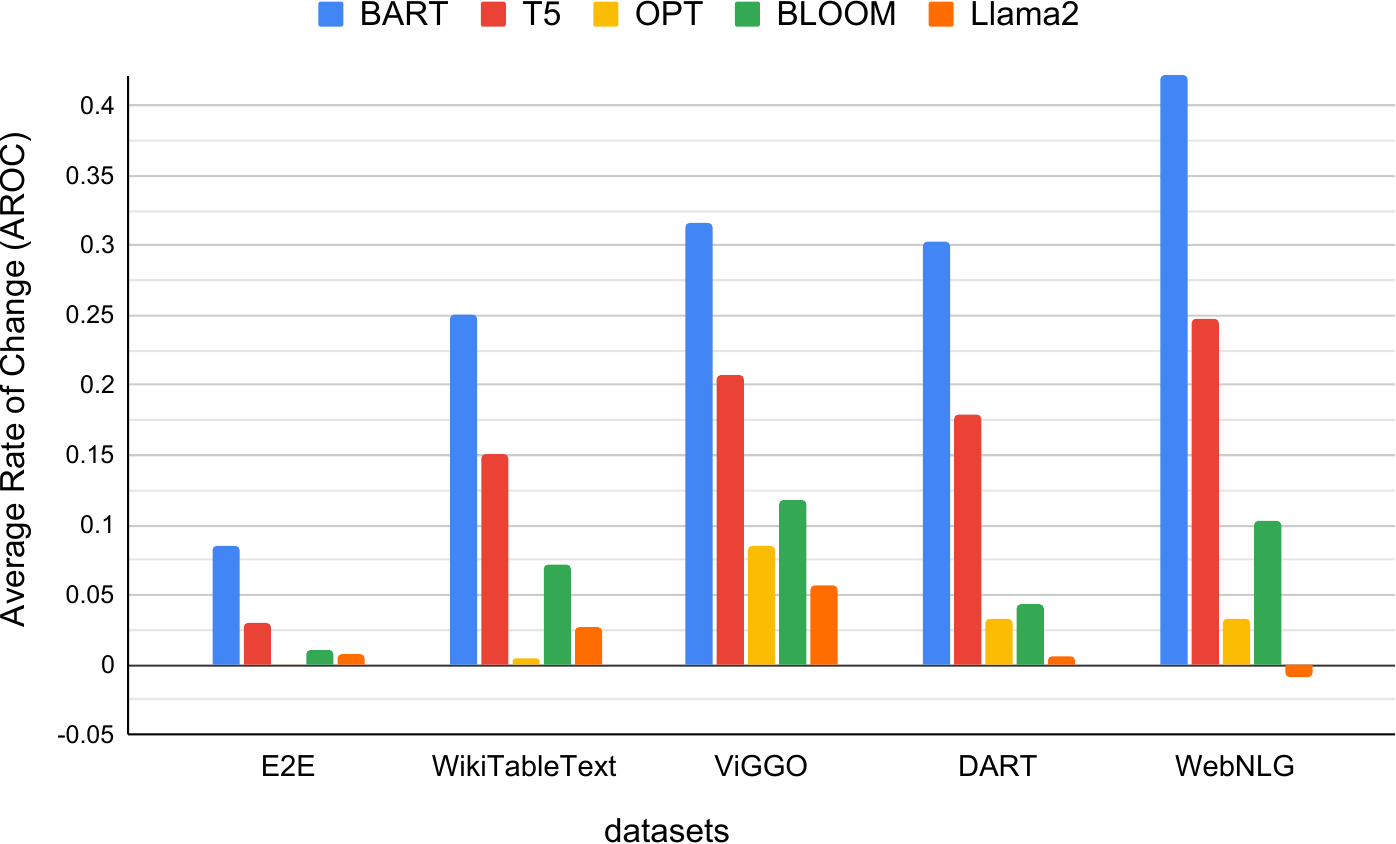}}
    \caption{AROC values for all LLM families across the five DTG datasets based on \textsc{QAFactEval} scores.
    Higher positive AROC values reflect a stronger positive correlation between increases in model size and factual consistency.}
    \label{fig:qafacteval-aroc}
\end{figure}

\begin{figure}[h]
    \centering
    \resizebox{\linewidth}{!}{
    \includegraphics[]{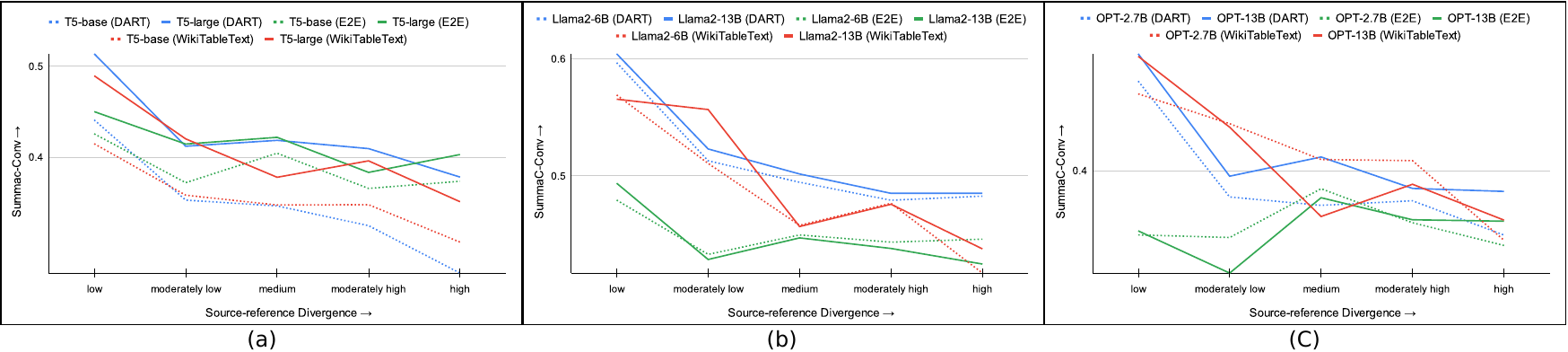}}
    \caption{Impact of source-reference divergence on factual consistency (measured through  \textsc{SummaC-Conv}) at time of inference on three popular DTG datasets (DART, E2E and WikiTableText). 
    The three sub-figures (a, b and c) represent results from three different families (T5, Llama2 and OPT), where small-sized LLMs (\textsf{T5-base}, \textsf{Llama2-6B} and \textsf{OPT-2.7B}) are represented with dotted line and large-sized LLMs (\textsf{T5-large}, \textsf{Llama2-13B} and \textsf{OPT-13B}) are represented with solid line.}
    \label{fig:divergence-summac-conv}
\end{figure}

\begin{figure}[h]
    \centering
    \resizebox{\linewidth}{!}{
    \includegraphics[]{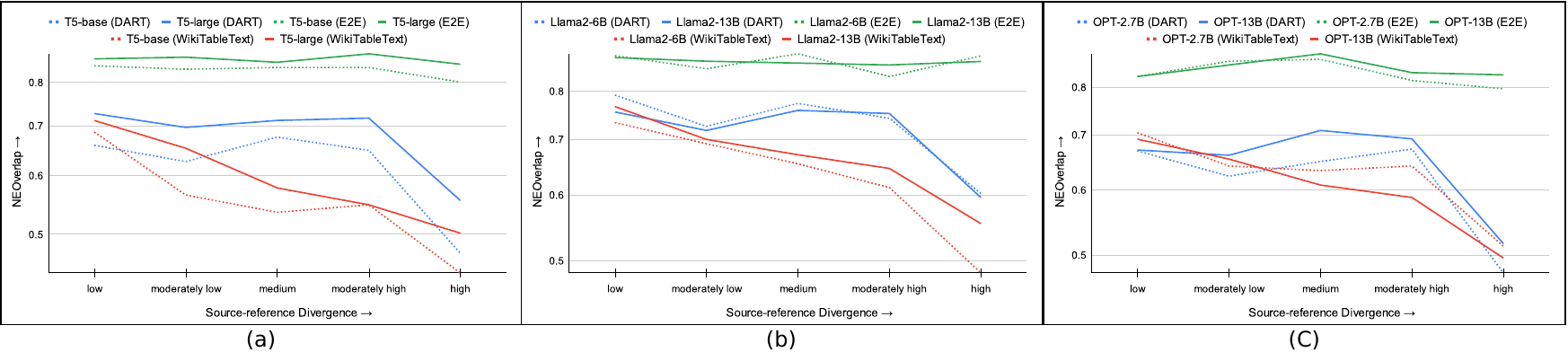}}
    \caption{Impact of source-reference divergence on factual consistency (measured through  \textsc{NEOverlap}) at time of inference on three popular DTG datasets (DART, E2E and WikiTableText). 
    The three sub-figures (a, b and c) represent results from three different families (T5, Llama2 and OPT), where small-sized LLMs (\textsf{T5-base}, \textsf{Llama2-6B} and \textsf{OPT-2.7B}) are represented with dotted line and large-sized LLMs (\textsf{T5-large}, \textsf{Llama2-13B} and \textsf{OPT-13B}) are represented with solid line.}
    \label{fig:divergence-neoverlap}
\end{figure}

\subsection{Results of Factual Consistency Evaluation Using \textsc{QAFactEval}}
\label{subsec:qafacteval}
\textsc{QAFactEval}~\cite{fabbri2022qafacteval} is a factual consistency metric that utilizes question generation and answering (QGA).
It comprises four key components: 1) Answer Selection: Noun phrase chunks are extracted as answers from the generated text; 2) Question Generation: The \textsf{BART-large} model generates questions based on these answers; 3) Question Answering: The \textsf{Electra-large} model answers the questions using the source text; 4) Answer Overlap Determination: The overlap is assessed using the LERC metric, which provides a score from 1 to 5. 
\textsc{QAFactEval} surpasses existing QGA-based metrics and complements NLI-based evaluations, with higher scores indicating greater factual consistency, reaching a maximum of 5.

\autoref{tab:qafacteval-numerical} presents the factual consistency scores evaluated using \textsc{QAFactEval} across all LLM families and DTG datasets. 
~\autoref{fig:qafacteval-aroc} illustrates the AROC values of \textsc{QAFactEval}, derived from \autoref{tab:qafacteval-numerical}, for five LLM families across all DTG datasets.

\subsection{Discussion of Factual Consistency Results}
From the factual consistency results across the four metrics—\textsc{SummaC-Conv}, \textsc{NEOverlap}, \textsc{AlignScore}, and \textsc{QAFactEval} (Tables~\ref{tab:summac-conv-numerical},~\ref{tab:neoverlap-numerical},~\ref{tab:alignscore-numerical}, and~\ref{tab:qafacteval-numerical})—we observe several key findings.
The Llama2 family consistently outperforms other models in factual consistency across all four metrics and nearly all DTG datasets.
However, smaller LLMs like \textsf{T5-large} and \textsf{BART-large} remain competitive, often outperforming larger families such as BLOOM and OPT on datasets like E2E (where lexical richness, as measured by TTR in~\autoref{tab:dataset_statistics}, is low, and dataset size is large) and DART (which has a large number of instances, as shown in~\autoref{tab:dataset_statistics}).
This indicates that smaller LLM families can excel depending on dataset characteristics. Larger LLM families like Llama2, BLOOM, and OPT tend to perform better on datasets such as WikiTableText and ViGGo, likely due to their higher lexical richness (see~\autoref{tab:dataset_statistics}). 
Similarly, for the WebNLG dataset, T5 and BART produce results similar to those of BLOOM and OPT, likely because WebNLG is a large dataset with moderate lexical richness.
Hence, while larger LLMs often exhibit better factual consistency than smaller ones, the characteristics of D2T datasets, particularly their size and lexical richness, are essential in shaping this performance.

AROC values across all four metrics (Figures~\ref{fig:summac-conv-aroc},~\ref{fig:neoverlap-aroc},~\ref{fig:alignscore-aroc}, and~\ref{fig:qafacteval-aroc}) show that increasing model size generally improves the factual consistency of LLM families across all DTG datasets. 
The T5 and BLOOM families exhibit significant improvements in factual consistency with larger models. 
In contrast, the Llama2 family shows less noticeable gains, and in some cases, a decline in performance as model size increases. 
We hypothesize that this may be due to the log-scale applied to the AROC values, where smaller LLM families (like T5 and BART) show a higher relative improvement in factual consistency compared to larger families (like Llama2). 
Overall, the AROC values across all four metrics clearly indicate that, for most LLM families, larger model sizes are correlated with improved factual consistency.

\begin{figure}[h]
    \centering
    \resizebox{\linewidth}{!}{
    \includegraphics[]{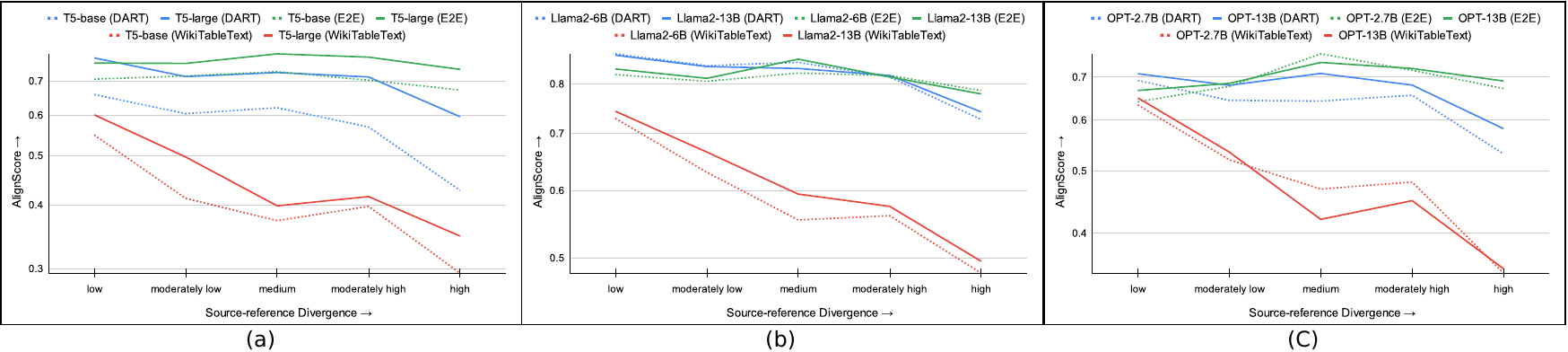}}
    \caption{Impact of source-reference divergence on factual consistency (measured through  \textsc{AlignScore}) at time of inference on three popular DTG datasets (DART, E2E and WikiTableText). 
    The three sub-figures (a, b and c) represent results from three different families (T5, Llama2 and OPT), where small-sized LLMs (\textsf{T5-base}, \textsf{Llama2-6B} and \textsf{OPT-2.7B}) are represented with dotted line and large-sized LLMs (\textsf{T5-large}, \textsf{Llama2-13B} and \textsf{OPT-13B}) are represented with solid line.}
    \label{fig:divergence-alignscore}
\end{figure}

\begin{figure}[h]
    \centering
    \resizebox{\linewidth}{!}{
    \includegraphics[]{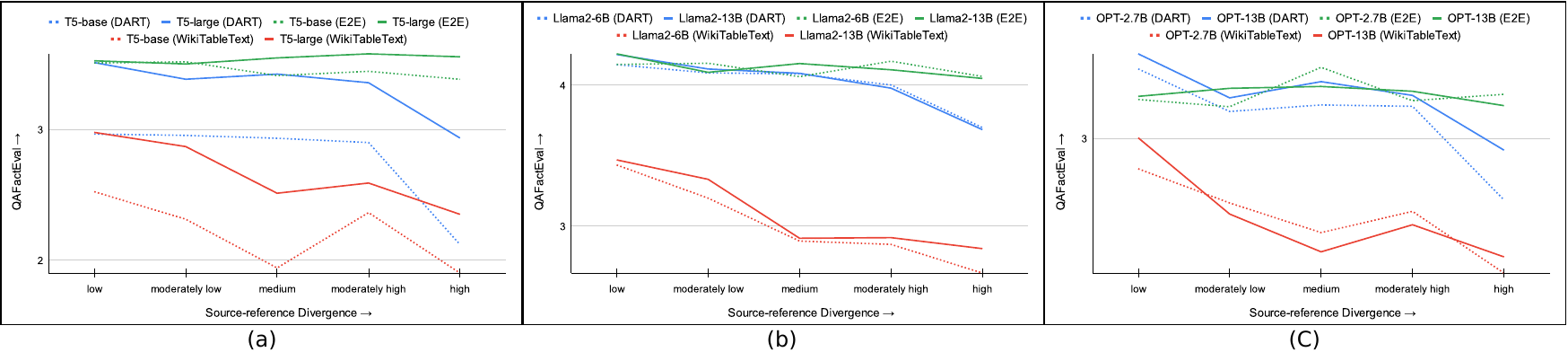}}
    \caption{Impact of source-reference divergence on factual consistency (measured through  \textsc{QAFactEval}) at time of inference on three popular DTG datasets (DART, E2E and WikiTableText). 
    The three sub-figures (a, b and c) represent results from three different families (T5, Llama2 and OPT), where small-sized LLMs (\textsf{T5-base}, \textsf{Llama2-6B} and \textsf{OPT-2.7B}) are represented with dotted line and large-sized LLMs (\textsf{T5-large}, \textsf{Llama2-13B} and \textsf{OPT-13B}) are represented with solid line.}
    \label{fig:divergence-qafacteval}
\end{figure}

\section{Impact of Source-reference Divergence on Factual Consistency}
\label{sec:results-divergence}
This section investigates how source-reference divergence impacts the factual consistency of LLMs in DTG tasks during the inference phase (i.e., at the time of text generation).
For this analysis, we focus on three LLM families (T5, OPT, and Llama 2) with six specific models: \textsf{T5-base}, \textsf{T5-large}, \textsf{OPT-2.7B}, \textsf{OPT-13B}, \textsf{Llama2-7B}, and \textsf{Llama2-13B}.
We evaluate the factual consistency of these LLMs using the four automatic metrics: \textsc{SummaC-Conv}, \textsc{NEOverlap}, \textsc{AlignScore}, and \textsc{QAFactEval}. 
To cover the three main types of DTG tasks, we select one dataset for each: E2E for MR-to-Text, WikiTableText for Table-to-Text, and DART for Graph-to-Text.
To better illustrate the effect of source-reference divergence, we divide the test/inference sets of each dataset into five groups based on their average source-reference divergence: \texttt{low}, \texttt{moderately low}, \texttt{medium}, \texttt{moderately high}, and \texttt{high}.
The \texttt{low} group contains instances with minimal divergence, while the \texttt{high} group includes instances with significant divergence.
Given that the source data ($s$) in DTG tasks is often semi-structured and non-textual, traditional methods of calculating source-reference divergence, such as higher-order n-gram matching or contextual representation, may fail to capture critical information.
Two common strategies for assessing source-reference divergence are unigram-based and longest common subsequence (LCS)-based approaches~\cite{dhingra2019handling}.
We choose the LCS-based approach over the unigram-based method because it often yields better performance by flexibly matching n-grams and capturing skip context.
Source-reference divergence ($\text{div}(s, r)$) is then defined based on the differences between the source text ($s$) and the reference text ($r$) at the unigram level, as follows:

\begin{align*}
    \text{div} (s, r) = 1 - \frac{1}{\operatorname{max}(|s|, |r|)}\operatorname{LCS}(s, r)
\end{align*}

Where $\vert x \vert$ denotes the cardinality, and $\operatorname{LCS}(x, y)$ represents the length of the longest common subsequence between $x$ and $y$ (both $x$ and $y$ are segmented by whitespaces).

From the results shown in Figures~\ref{fig:divergence-summac-conv},~\ref{fig:divergence-neoverlap},~\ref{fig:divergence-alignscore}, and~\ref{fig:divergence-qafacteval}, we observe two important details regarding factual consistency in the presence of source-reference divergence.
First, the factual consistency of all LLMs, regardless of model family or size, decreases with an increase in source-reference divergence. 
However, in the E2E dataset, the decrease in factual consistency with source-reference divergence is lower compared to the DART and WikiTableText datasets.
We believe this is due to the higher lexical richness of the DART and WikiTableText datasets compared to the E2E dataset.
Second, it is evident that within each LLM family, larger models consistently outperform smaller ones in terms of factual consistency.
This suggests that increasing the model size of LLMs makes them more agnostic to source-reference divergence.
This finding indicates that choosing larger model-sized LLMs is beneficial when source-reference divergence is present in DTG tasks.

\section{Human Evaluation}
\label{sec:human-evaluation}
Text generation tasks like DTG require human evaluation due to the complexity of natural language, and human assessment is often regarded as the gold standard for factual consistency~\cite{reiter2018structured, chaganty2018price}.
For our human evaluation of factual consistency, we selected three LLM families (T5, OPT, and Llama 2), each with two models (\textsf{T5-base}, \textsf{T5-large}, \textsf{OPT-2.7B}, \textsf{OPT-13B}, \textsf{Llama2-7B}, and \textsf{Llama2-13B}), and three DTG datasets: E2E, WikiTableText, and DART.
Three university graduate students, selected from different academic disciplines to reduce bias, were chosen to annotate a sample of 20 generated texts for each combination of LLM and dataset.
Each annotator was instructed to categorize factual consistency based on the presence of incorrect or irrelevant facts with respect to the given source data.
Responses were collected as `yes'/`no' answers from each annotator. 
Finally, we aggregated the results using a majority voting method and expressed the results as percentages.
The inter-annotator agreement in our experiments, measured by Krippendorff's alpha ($\alpha$), was 0.76, indicating moderate to high agreement.

\begin{table}[h]
    \centering
    \caption{Human evaluation of factual consistency.
    A lower value indicates higher factual consistency.}
    \label{tab:human-evaluation}
    \resizebox{0.4\linewidth}{!}{
    \begin{tabular}{@{}ccccc@{}}
        \toprule
        family & model & E2E & DART & WikiTableText \\ \midrule
        \multirow{2}{*}{T5} & T5-base & 23\% & 32\% & 37\% \\
         & T5-large & 15\% & 22\% & 32\% \\ \midrule
        \multirow{2}{*}{OPT} & OPT-2.7B & 19\% & 25\% & 18\% \\
         & OPT-13B & 16\% & 22\% & 14\% \\ \midrule
        \multirow{2}{*}{Llama 2} & Llama2-7B & 12\% & 15\% & 21\% \\
         & Llama2-13B & 12\% & 17\% & 17\% \\ \bottomrule
    \end{tabular}}
\end{table}

\autoref{tab:human-evaluation} shows the human evaluation results for factual consistency.
We can see that, in human evaluation as well, the Llama 2 family performs better than OPT and T5. 
Similar to the automatic metric results, we find that T5 performs similarly to the OPT family on the DART and E2E datasets. 
However, OPT outperforms T5 on the WikiTableText dataset. 
The similarity between the results of human evaluation and automatic metric-based evaluations further supports the overall findings of this paper.

\section{Conclusion}
\label{sec:conclusion}
This paper presents a comprehensive evaluation of the factual consistency of large language models (LLMs) in data-to-text generation (DTG) tasks.
Through extensive experiments with five leading LLM families across five popular DTG datasets, we utilize four state-of-the-art automatic metrics—\textsc{SummaC-Conv}, \textsc{QAFactEval}, \textsc{AlignScore}, and \textsc{NEOverlap}—alongside crucial human evaluations.
Our findings reveal three key insights. 
First, Llama 2 consistently excels in generating factually consistent text. 
However, smaller models like T5 and BART can also achieve strong factual consistency, especially on larger datasets with lower lexical diversity.
Second, the average rate of change (AROC) analysis demonstrates that increasing model size (number of model trainable parameters) generally improves factual consistency in DTG tasks.
Third, we observe that source-reference divergence—where the reference text deviates semantically from the source—typically reduces the factual consistency of LLMs in DTG. 
We believe that these key findings, along with our detailed evaluation, will serve as a valuable foundation for researchers and practitioners seeking to effectively deploy LLMs in DTG tasks where factual consistency is paramount.
While this paper focuses on parameter-efficient fine-tuned LLMs (via QLoRA) for DTG, future work will explore other parameter-efficient fine-tuning approaches, such as prompting.

\section*{Acknowledgment}
This research is partially supported by the Indo-French Centre for the Promotion of Advanced Research (IFCPAR/CEFIPRA) through CSRP Project No. 6702-2.

\bibliography{references/final_base} 
\bibliographystyle{plainnat}
\end{document}